\documentclass[letterpaper]{article}
\usepackage[preprint]{aaai2027}

\usepackage[hyphens]{url}
\usepackage{graphicx}
\usepackage{natbib}
\usepackage{caption}
\usepackage{algorithm}
\usepackage{algorithmic}
\newcommand{\Comment}[1]{\hfill\textit{// #1}}
\usepackage{booktabs}
\usepackage{subcaption}
\usepackage{multirow}
\usepackage{colortbl}
\definecolor{graybg}{gray}{0.90}
\newcommand{\gref}[1]{(#1)}

\usepackage{amsmath,amssymb}

\newcommand{\N}{\mathcal{N}}

\newcommand{\eps}{\varepsilon}
\newcommand{\xzero}{x_0}

\title{ReNFT: Repairing Mode Collapse in Reward Post-Training via Internal Probability-Mass Recalibration}

\author{
  Yuchen Bao\textsuperscript{\rm 1, 2},
  Chao Wen\textsuperscript{\rm 2},
  Haowei Wang\textsuperscript{\rm 2},
  Ruoxin Chen\textsuperscript{\rm 2},
  Donghao Luo\textsuperscript{\rm 2},
  Jiahui Zhan\textsuperscript{\rm 2},
  Wenjian Huang\textsuperscript{\rm 1},
  Shen Chen\textsuperscript{\rm 2},
  Yiting Wang\textsuperscript{\rm 2},
  Taiping Yao\textsuperscript{\rm 2},
  Chengjie Wang\textsuperscript{\rm 2},
  Shouhong Ding\textsuperscript{\rm 2},
  Jianguo Zhang\textsuperscript{\rm 1}\thanks{Corresponding author.}
}
\affiliations{
  \textsuperscript{\rm 1}Southern University of Science and Technology\\
  \textsuperscript{\rm 2}Tencent Youtu Lab
}

\begin{document}
\maketitle

\begin{abstract}
Reward post-training of diffusion generators inevitably concentrates probability mass on a few reward-favored modes, a mode collapse that erases within-prompt diversity. Existing methods for mitigating collapse rely on external signals or interfaces, augmenting the reward with perceptual objectives, adjusting reference regularization, or modifying the text encoder, but none repairs an adapter that has already collapsed while preserving the acquired reward. We observe that online post-training primarily reallocates probability mass over capabilities inherited from pretraining rather than learning new visual content. Collapse is therefore suppression, not deletion, and can be reversed from within the generator. We propose \textbf{ReNFT}, which repairs a high-reward, low-diversity adapter through internal probability-mass recalibration. Unconditional probes first prioritize ``anti-hub'' prompts where the prompt-independent bias is easiest to expose. Two policy-dominated mixed routes then generate matched counterfactual proposals from the same prompt and initial noise, one probing the frozen base direction for suppressed alternatives and the other exposing the post-trained unconditional tendency. Reward ranking with an adaptive flipping guard assigns pull and push roles, and a joint-and-paired NFT update realizes the repair. On PickScore and GenEval, \textbf{ReNFT} retains 98.9\% and 99.0\% of NFT's reward while improving DreamSim-Div by 58.8\% and 55.0\%, respectively, offering a complementary alternative to external interventions.
\end{abstract}

\section{Introduction}

Reward-based post-training has become a practical way to align diffusion and
flow generators beyond pretraining. Preference optimization and
policy-gradient methods, represented by DiffusionDPO and Flow-GRPO, improve
alignment by increasing the probability of reward-preferred
outputs~\cite{wallace2024diffusiondpo,liu2025flowgrpo}. DiffusionNFT
(Negative-aware Fine-Tuning, NFT)~\cite{zheng2025diffusionnft} makes this
alignment markedly more efficient: it replaces likelihood-ratio policy
gradients with forward-process reconstruction, requires no classifier-free
guidance, and converges much faster. The same feedback loop, however,
inevitably concentrates probability mass; the more efficient the
optimization, the more severe the concentration: whenever a narrow visual
mode repeatedly receives high reward, different initial noises under the
same prompt are mapped to similar outputs. Reward differences
among such semantically equivalent samples encode the preference of the
reward model rather than genuine quality, so optimizing them only sharpens
the preference. The result is a high-reward adapter whose within-prompt
structural and stylistic diversity has collapsed, most severely under the
fastest optimizer, NFT.

\begin{figure*}[!t]
  \centering
  \includegraphics[width=0.8\textwidth]{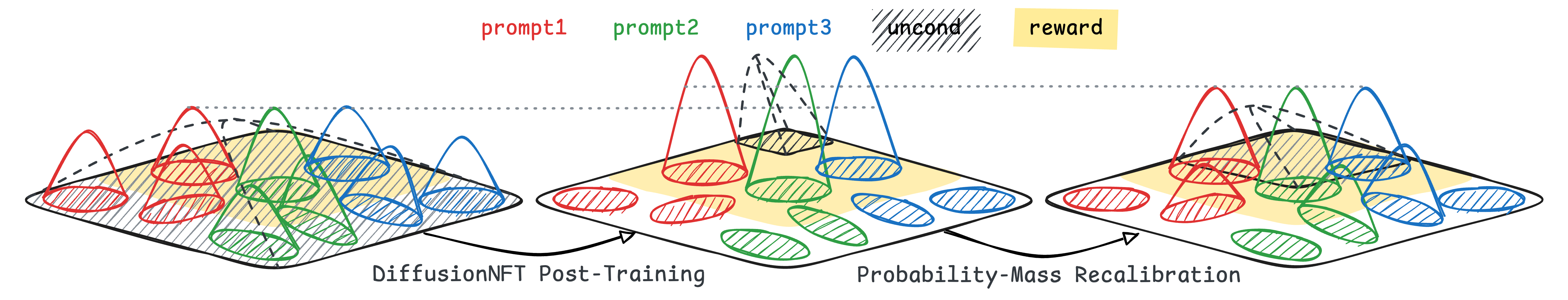}
  \caption{\textbf{Conceptual view of probability-mass recalibration.}
  Pretraining spreads probability across the modes of each prompt, while
  unconditional behavior spans their marginal range. NFT collapses each prompt
  onto a reward-near mode and unconditional behavior onto a fixed mode,
  suppressing alternatives. \textbf{ReNFT} restores unconditional coverage and
  reward-boundary modes while leaving low-quality modes outside suppressed.
  Gray dashed lines compare the prompts' highest-reward mode probabilities.}
  \label{fig:distribution-recalibration}
\end{figure*}

Existing methods for mitigating collapse mainly intervene at three positions. Reward-shaping
approaches augment the reward with external perceptual
signals~\cite{liu2025diversegrpo,tan2026pec,liu2025drift}, but can only
reweight among samples the policy still generates. Regularization-based approaches align the policy toward the base
model~\cite{liu2025diversegrpo,he2025gardo}, tying achievable reward to the
base ceiling. Interface-level methods modify text
representations~\cite{hu2026e2po,chen2025d2align}, but depend on the
architecture of the reward model. We therefore ask:
\emph{can an already-collapsed adapter be repaired using only distributions
already inside the generator, without external diversity objectives or
text-encoder modification, and without sacrificing its acquired reward?}

Our answer begins with a distinction: pretraining learns from external
images, whereas online post-training supervises the generator with its own
scored samples, so it reallocates probability mass over capabilities
inherited from the base model, amplifying reward-favored modes while
starving alternatives. Collapse is therefore
suppression, not deletion: suppressed modes remain reachable through internal
routes of the generator, and restoring their probability mass requires
no new visual knowledge. We refer to this operational repair view as
\emph{probability-mass recalibration}.

Recalibration first requires an internal signal that exposes where the mass
has gone. The post-trained unconditional route provides such a readout: it
shares the updated parameters but drops the prompt condition, exposing what
the adapter injects without being asked. This readout is diagnostic rather
than a training target: the repair must act on conditional generations,
and optimizing the unconditional direction directly offers no handle on
per-prompt structure. We therefore use it in two indirect ways: locating
prompts far from the exposed tendency, and constructing counterfactual
samples inside ordinary conditional trajectories, where the tendency
becomes testable by the original reward model.

These observations motivate \textbf{ReNFT} (\emph{Repair NFT}), which repairs high-reward,
low-diversity adapters through internal probability-mass recalibration.
To expose the bias where it is
most distinguishable, unconditional probes first prioritize
\emph{``anti-hub''} prompts, those farthest from the unconditional
\emph{hub}, the shared mode toward which the unconditional distribution
concentrates under reward hacking. To obtain candidates without any external signal, two
policy-dominated mixed routes generate matched proposals from the same prompt
and initial noise: one probes the frozen base direction for suppressed
alternatives, while the other exposes the post-trained unconditional
tendency. The original reward then ranks the pair into pull and mirrored
push targets, and an adaptive flipping guard keeps at least half of the
pull targets on the base-probing route.
A native-noise joint update then anchors the preference at the
shared trajectory origin, and fresh-noise paired updates propagate it across
intermediate noise levels. A frozen image--text encoder is used only to
prioritize prompts; it never enters the reward, advantage, or repair loss.

\noindent In summary, our contributions are as follows:
\begin{itemize}
  \item We formulate reward-induced mode collapse as an internal
        probability-mass reallocation problem and study the repair of a
        severely reward-hacked, high-reward adapter rather than
        diversity-aware training from scratch.
  \item We identify the post-trained unconditional route as a free internal
        readout of reward bias and exploit it in two ways: prioritizing
        \emph{``anti-hub''} prompts farthest from the exposed tendency, and
        injecting the tendency into conditional trajectories to turn a hidden
        bias into an explicit, reward-verifiable candidate.
  \item We realize probability-mass recalibration through two internal routes
        of the same generator: branched from the same prompt and initial noise, the
        frozen base route expands probability mass back toward suppressed
        alternatives, while the unconditional route exposes the learned bias
in a mixed endpoint as a mirrored-push candidate; pull and
push follow the reward gap of each pair rather than route identity. This
        pair construction is training-only;
        the repaired adapter samples with the standard conditional forward
        pass, preserving 98.9--99.0\% of NFT's
        reward while improving DreamSim-Div~\cite{fu2023dreamsim} by
        58.8\% and 55.0\% on PickScore~\cite{pickscore} and
        GenEval~\cite{geneval}, respectively.
\end{itemize}


\section{Related Work}
\label{sec:related}

\begin{figure*}[!htb]
  \centering
  \includegraphics[width=0.85\textwidth]{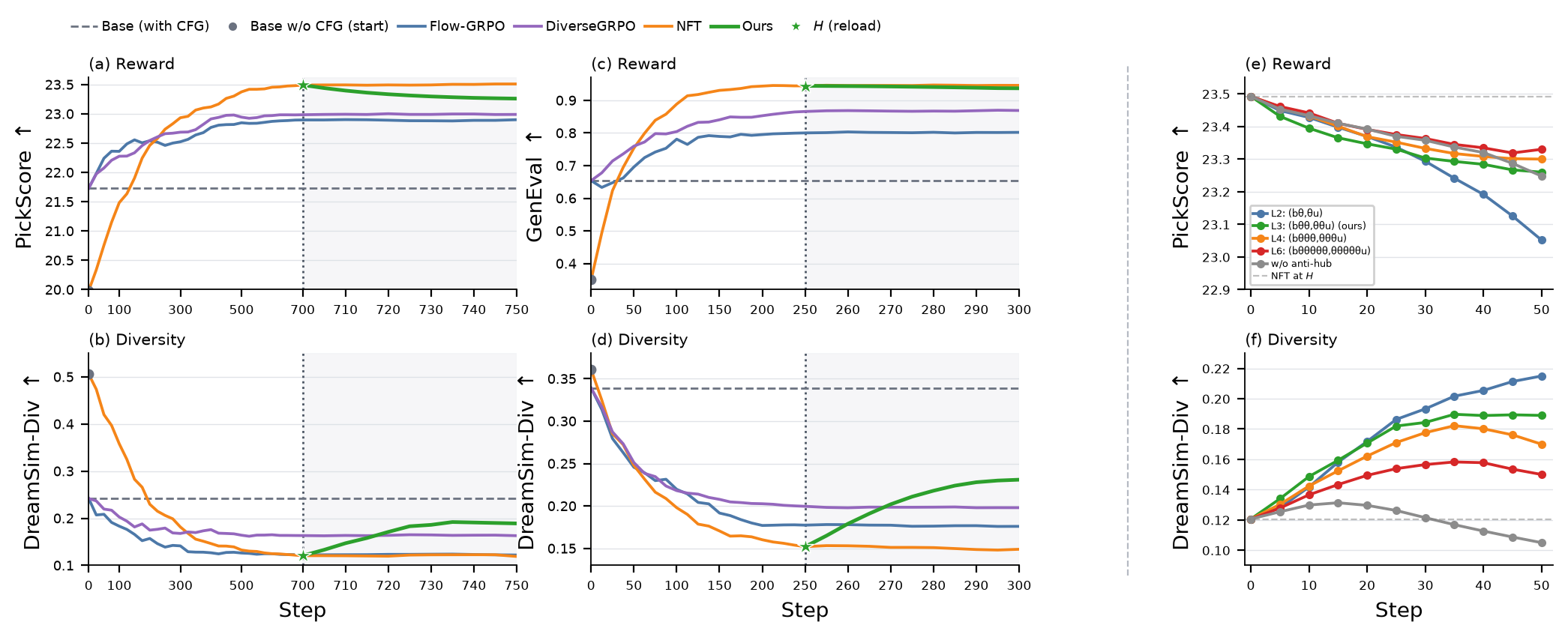}
  \caption{\textbf{Training dynamics and ablations.}
  (a)(b): PickScore reward and DreamSim-Div;
  (c)(d): GenEval reward and DreamSim-Div.
  \textbf{ReNFT} branches from the hacked checkpoint $H$ (star) and repairs for
  50 steps; the dashed line marks the base generator.
  (e)(f): mixed-route pattern and anti-hub ablations after 50 repair steps
  from the same $H$; the dashed line marks NFT at $H$.}
  \label{fig:training-dynamics}
\end{figure*}

DiffusionDPO~\cite{wallace2024diffusiondpo} adapts preference optimization to
reward post-training of diffusion and flow generators;
Flow-GRPO~\cite{liu2025flowgrpo}, DenseGRPO~\cite{deng2026densegrpo}, and
AWM~\cite{xue2025awm} improve group-relative optimization, reward density, or
stability; and DiffusionNFT~\cite{zheng2025diffusionnft} transfers supervision
to forward-process reconstruction; \citet{surveygrpo} survey this line.
Subsequent variants refine clipping, credit assignment, self-correction, and
distillation~\cite{ping2026flowdppo,tong2026tpgrpo,qin2026soar,li2026diffusionopd,fang2026flowopd,go2026stitchvm,li2025uniworldv2}.
Without a mechanism preserving broad support, repeatedly
reinforcing self-generated high-reward samples concentrates probability mass
on a few modes, i.e., mode collapse, as also observed for on-policy,
reverse-divergence objectives~\cite{zheng2026causalrcm}.
Interventions differ mainly in where they inject the diversity signal.

\noindent\textbf{Reward-shaping approaches.}
A first line augments the reward or advantage with perceptual diversity
signals from external models. DiverseGRPO~\cite{liu2025diversegrpo} clusters
same-prompt samples in CLIP space and adds an exploration bonus inversely
proportional to cluster size. PEC~\cite{tan2026pec} replaces policy entropy
with a perceptual-entropy proxy from rollout states. DRIFT~\cite{liu2025drift}
combines reward-concentrated rollout selection, prompt perturbation, and
potential-based shaping. Related work replaces the reward signal itself with
pairwise win rates, gated or adversarial advantages, or set-level
statistics~\cite{wang2025prefgrpo,mao2025advgrpo,li2026distributionwise}.

\noindent\textbf{Regularization-based approaches.}
A second line modifies reference regularization.
DiverseGRPO~\cite{liu2025diversegrpo} applies stronger KL regularization
during early denoising steps and relaxes it later.
GARDO~\cite{he2025gardo} gates penalization to high-uncertainty samples,
periodically updates the EMA reference, and amplifies rewards for
high-quality diverse samples.

\noindent\textbf{Interface-level approaches.}
A third line modifies the semantic interface through which prompts or rewards
enter optimization. D$^2$-Align~\cite{chen2025d2align} learns a directional
correction in the text-embedding space of a frozen reward model.
E$^2$PO~\cite{hu2026e2po} perturbs content-token embeddings on the generator
side and anneals the perturbation during denoising.

\textbf{ReNFT} is complementary to all three lines: its repair signal comes from
components already inside the post-trained generator (a frozen encoder serves
only to prioritize prompts). A natural question is
whether these methods can also \emph{repair} an already-collapsed adapter:
reward-shaping and interface-level methods are designed for prevention during
training, while KL-style anchoring can restore suppressed modes but ties the
attainable reward to the base model. \textbf{ReNFT} realizes this repair through
two internal routes of the same generator: a frozen base route that probes
suppressed alternatives and an unconditional route that exposes
prompt-independent bias. Trajectory-routing
work~\cite{soboleva2025tlora,cao2026dynamicfusion,yin2026fera,jin2026stagewise}
likewise shows that route identity and timestep matter, but does not construct
reward-ranked, shared-noise counterfactuals.
\section{Preliminaries}
\label{sec:prelim}

\begin{figure*}[!t]
  \centering
  \includegraphics[width=1.0\textwidth]{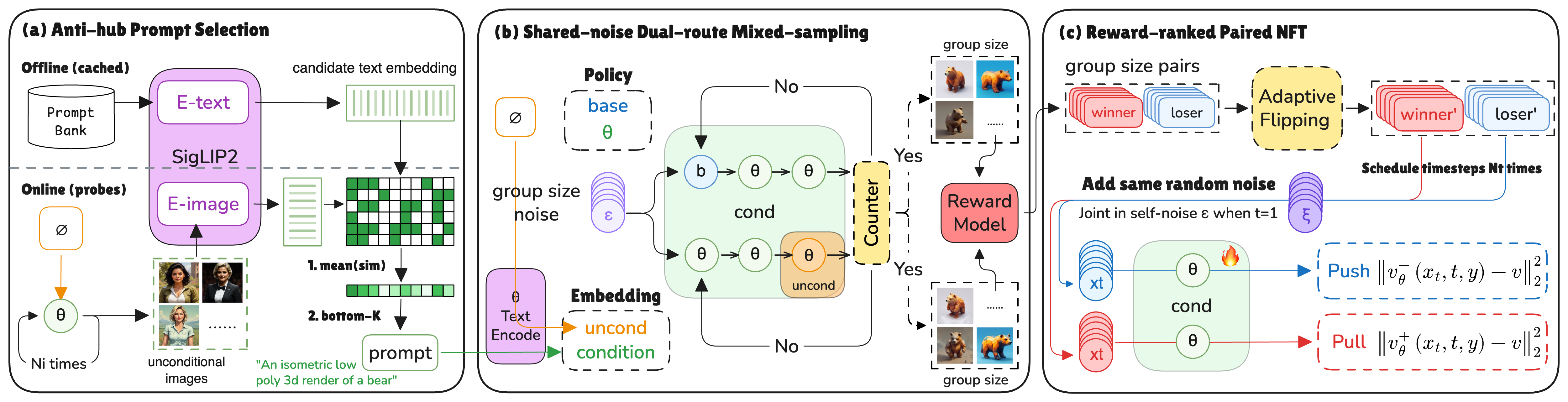}
  \caption{\textbf{ReNFT pipeline.} (a) Unconditional probes prioritize
  anti-hub prompts; (b) two policy-dominated mixed routes generate matched
  proposals from the same prompt and initial noise; (c) reward ranking and the
  adaptive flipping guard assign pull/push targets for joint-and-paired NFT
  repair. The VAE decoder is omitted for clarity.}
  \label{fig:framework}
\end{figure*}

We introduce the flow-matching notation and the NFT-style forward-process
post-training update that form the basis of our method.

\subsection{Flow-Matching Generation}
\label{sec:prelim-flow}
Let $y$ denote a text prompt, $\xzero$ a clean latent, and
$\eps\sim\N(0,I)$ the initial noise. Under rectified-flow interpolation,
\begin{equation}
  x_t = (1-t)\xzero + t\eps, \quad t\in[0,1],
  \label{eq:flow-interp}
\end{equation}
the target transport velocity is $v=\eps-\xzero$, and the model predicts
$v_\theta(x_t,t,y)$ trained with the standard flow-matching regression
objective. We denote three routes through this sampler: $b$ (frozen base),
$\theta$ (conditional policy), and $u$ (unconditional route of the current
policy); an EMA-smoothed copy of the policy provides a stable reference,
with its velocity written $v_{\mathrm{old}}$.

\subsection{NFT-Style Forward-Process Post-Training}
\label{sec:prelim-nft}
Reward-based post-training reshapes the distribution of a pretrained generator
using a reward model $R(\xzero,y)$. DiffusionNFT~\cite{zheng2025diffusionnft}
avoids likelihood-ratio policy gradients by applying reward supervision
through forward-process regression. Each generated endpoint $\xzero$ (from
a rollout of $N_i$ ODE steps) is re-noised at $N_t$ training timesteps to
$x_t=(1-\sigma_t)\xzero+\sigma_t\xi$, with fresh $\xi\sim\N(0,I)$ and noise
level $\sigma_t{=}t$, and the model regresses the target velocity
$v=\xi-\xzero$. High-reward endpoints are reconstructed through the
ordinary prediction
\begin{equation}
  v_\theta^{+}(x_t,t,y)=v_\theta(x_t,t,y),
  \label{eq:nft-vplus}
\end{equation}
while low-reward endpoints are pushed away through a mirrored prediction
around the EMA reference. With the official default $\beta=1$ in
$(1+\beta)v_{\mathrm{old}}-\beta v_\theta$, this reduces to
\begin{equation}
  v_\theta^{-}(x_t,t,y)=2\,v_{\mathrm{old}}(x_t,t,y)-v_\theta(x_t,t,y).
  \label{eq:nft-vminus}
\end{equation}
The combined objective over a rollout group is
\begin{equation}
  \mathcal{L}_{\mathrm{NFT}}=
  \underbrace{\left\|v_\theta^{+}(x_t,t,y)-v\right\|_2^2}_{\text{pull}}
  +\underbrace{\left\|v_\theta^{-}(x_t,t,y)-v\right\|_2^2}_{\text{push}}.
  \label{eq:nft-loss}
\end{equation}
This MSE formulation trains quickly without likelihood ratios or SDE
rollouts, and the EMA reference in $v_\theta^{-}$ anchors the update and
stabilizes post-training (ablated in the appendix). Equivalently,
the loss can be written in endpoint space as $\ell(\hat{x}_0(v),\xzero)$
with $\hat{x}_0(v)=x_t-\sigma_t v$; our implementation uses this form with
a stop-gradient factor for loss-order adjustment. In our repair setting,
the pull and push branches will operate on \emph{different} endpoints
rather than the same $\xzero$; only at the pure-noise step $\sigma_t=1$
do the two trajectories share an identical state.

Online post-training supervises the model with its own sampled endpoints;
when reward concentrates in a narrow mode, this acts as self-distillation
that sharpens it. \textbf{ReNFT} retains the NFT parameterization but replaces
this one-sided supervision with a matched local comparison
(Section~\ref{sec:method}).

\section{Methodology}
\label{sec:method}

We address the repair of a high-reward, low-diversity adapter without adding
an external diversity objective or modifying text representations of the
generator. \textbf{ReNFT} instead constructs candidate endpoints from available routes of the
generator and ranks them with the same reward used for
post-training. Section~\ref{sec:repair-view} first explains why an
already-collapsed adapter can be repaired through these internal routes.
Figure~\ref{fig:framework} summarizes the ensuing pipeline: unconditional
probes prioritize anti-hub prompts where the prompt-independent bias is
easiest to expose (Section~\ref{sec:anti-hub-update}); two policy-dominated
mixed routes generate matched counterfactuals from the same prompt and initial
noise (Section~\ref{sec:dual-pattern}); and reward ranking assigns pull/push
targets, optimized by native-noise joint and fresh-noise paired NFT updates
(Section~\ref{sec:paired-forward}).

\subsection{Repair View: Post-Training Reweights an Inherited Distribution}
\label{sec:repair-view}
We study a repair setting in which an adapter $\theta_H$ already achieves high
reward but maps different initial noises under the same prompt to a narrow set
of outputs. The frozen base generator and $\theta_H$ share the same pretrained
support; reward adaptation changes which parts of that support are easy to
reach, but suppressed modes may remain accessible through routes closer to the
base model. Collapse is therefore probability-mass compression rather than
deletion, and repair should be possible from within.

A coarse mode-level view makes the feedback loop explicit. If $m$ indexes
a visual mode, reward-only adaptation can be heuristically written as
\begin{equation}
  p_{k+1}(m\mid y) \propto p_k(m\mid y)
  \exp\!\left(\eta\,\bar R_k(m,y)\right),
  \label{eq:mode-reweighting}
\end{equation}
where $\bar R_k(m,y)$ is the average reward of sampled outputs in mode $m$
and $\eta>0$ scales the reward-induced reweighting.
Equation~\eqref{eq:mode-reweighting} is an interpretation of this feedback,
not the exact update law. Figure~\ref{fig:distribution-recalibration}
illustrates the view: per-prompt modes contract into narrow spikes as the
loop reinforces the favored mode, then reopen during repair. Conditional
modes that had lost probability regain output mass, and the unconditional
distribution range expands accordingly.

This feedback also explains why re-optimizing reward cannot repair collapse:
among semantically equivalent samples, reward differences encode preference
rather than quality, so the reweighting only sharpens that preference. Repair
instead needs reward gaps that reflect genuine quality differences.
\textbf{ReNFT} achieves this by constructing matched counterfactuals from
different internal routes (Sections~\ref{sec:anti-hub-update}--\ref{sec:paired-forward}),
making the reward gap between paired endpoints informative rather than a
restatement of the model bias. We validate this reversibility in
Section~\ref{sec:exp}.

\subsection{Reading Post-Training Bias from the Unconditional Route}
\label{sec:anti-hub-update}
The unconditional route is the empty-prompt velocity
$u_\theta(x,t)=v_\theta(x,t,\varnothing)$. Although it shares $\theta$'s
updated parameters, it has no separate reward target; its changes therefore
expose prompt-independent side effects of conditional post-training: recurring
structures, styles, or textures the adapter injects even when not
requested, rather than directed optimization. Unconditional samples from a
fixed noise grid contract from a diverse base into a narrow hub after
post-training and progressively reopen during repair (see the appendix for
unconditional samples across training steps). We use these
tendencies as an internal bias probe, whose diagnostic
validity is evaluated rather than assumed.

We first use this readout to prioritize prompts on which the bias is easiest to
expose, as shown in Figure~\ref{fig:framework}a. Offline, the prompt bank is
encoded once by the SigLIP2 text encoder~\cite{tschannen2025siglip2} and
cached. Online, $M$
empty-prompt probes $\{u_j\}_{j=1}^{M}$ are generated from the current policy
and encoded by the SigLIP2 image encoder. Each candidate prompt $y$ is scored
by
\begin{equation}
  s(y)=\frac{1}{M}\sum_{j=1}^{M}
  \langle e_{\mathrm{text}}(y),e_{\mathrm{img}}(u_j)\rangle ,
  \label{eq:anti-hub-score}
\end{equation}
and the bottom-$K$ prompts (lowest mean similarity) are selected as
\emph{anti-hub prompts}, those farthest from the current unconditional
tendency, where the exposed bias is most likely to yield distinguishable
proposals. The frozen encoder serves only prompt prioritization; it never
enters the reward, advantage, or repair loss. The same route also serves
constructively in the mixed sampling of Section~\ref{sec:dual-pattern}, where
it exposes the bias inside conditional trajectories for reward ranking.

\subsection{Constructing Internal Counterfactual Proposals}
\label{sec:dual-pattern}
The frozen base route and the unconditional route play complementary roles. A
base step introduces a base-like vector-field direction and can propose
alternatives that are sampled less often by the post-trained policy; an
unconditional step exposes the prompt-independent bias identified in
Section~\ref{sec:anti-hub-update}. The conditional route $\theta$ occupies
most steps by design, limiting deviation from the high-reward policy.
Crucially, neither route is hard-coded as a pull or push branch: mixed
routing generates nearby alternatives, and the reward decides their roles
(Figure~\ref{fig:framework}b).

\begin{figure*}[!t]
  \centering
  \includegraphics[width=0.95\textwidth]{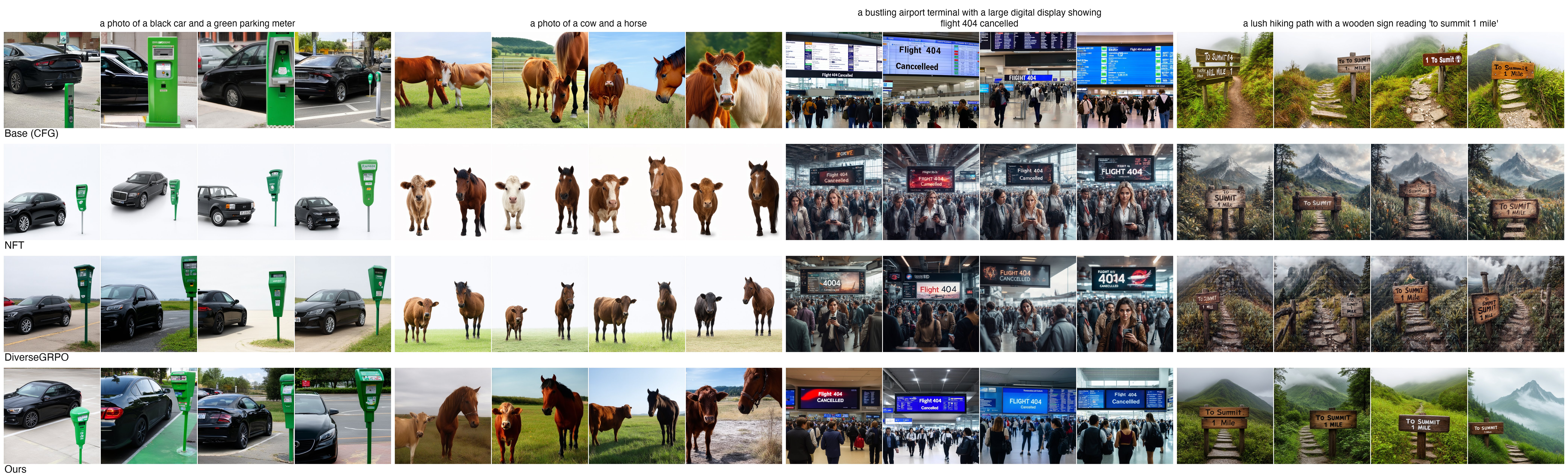}
  \caption{\textbf{Qualitative comparison on GenEval and OCR prompts.} Each
  prompt group shows four samples from the same method under different initial
  noises.}
  \label{fig:qual-geneval}
\end{figure*}

\begin{figure}[!t]
  \centering
  \includegraphics[width=0.9\linewidth]{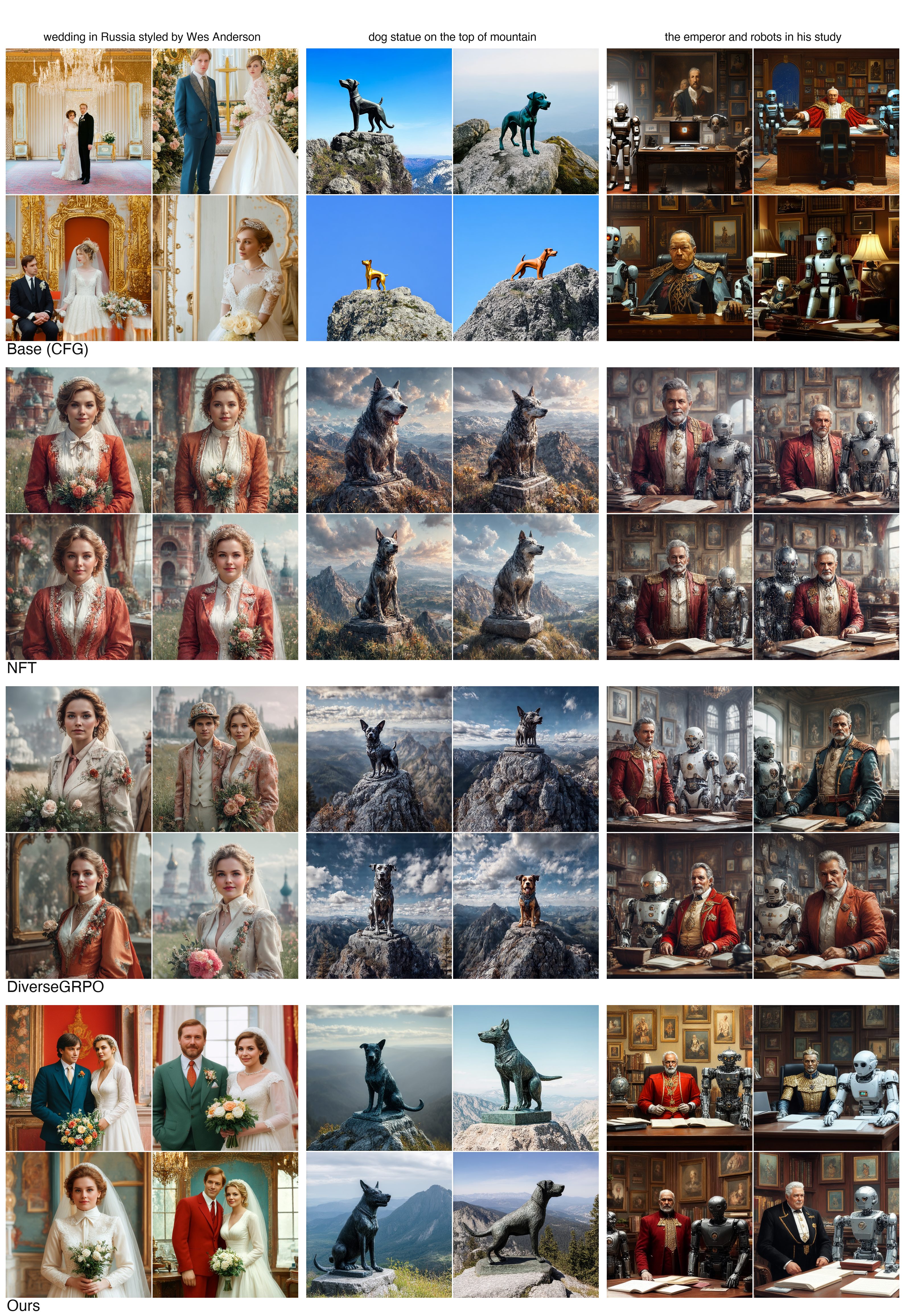}
  \caption{\textbf{Qualitative comparison on PickScore prompts.} Each prompt
  group shows four samples from the same method under different initial
  noises.}
  \label{fig:qual-pickscore}
\end{figure}

For a routing pattern $P$, let $T_{P,y}$ map initial noise to a terminal
sample. The final configuration repeats two three-step blocks across the
rollout,
\begin{equation}
  P_A=(b\theta\theta)\cdots, \qquad P_B=(\theta\theta u)\cdots.
  \label{eq:mixed-patterns}
\end{equation}
Both routes diverge from the same initial noise $\eps$, keeping two thirds of
each route conditional and avoiding long $bb$ or $uu$ blocks that stray too
far from the policy (cf.~stage-wise analyses~\cite{jin2026stagewise}).

For a selected prompt $y$ and shared initial noise $\eps$, the matched endpoints
are $x_0^A=T_{P_A,y}(\eps)$ and $x_0^B=T_{P_B,y}(\eps)$.
Sharing $(y,\eps)$ makes routing the main changed factor: two pure-$\theta$
trajectories would coincide, while a direct base-versus-policy comparison is
too one-sided; the two mixed routes instead perturb the policy in different
directions, and either may win for a particular $(y,\eps)$. Trajectory-level
branching has likewise been exploited for alignment and credit assignment in
TMPO and DenseGRPO~\cite{li2026tmpo,deng2026densegrpo}; we instead branch two
route mixtures from identical noise to construct a reward-testable
counterfactual pair.

We rank the endpoints with the original reward
(Figure~\ref{fig:framework}b, right) and obtain
$(x_0^H,x_0^L)$; when rewards differ, the higher-reward endpoint is always
the pull target and the lower the push target, regardless of route. The
matched reward gap
\begin{equation}
  \Delta R(y,\eps)=R(x_0^A,y)-R(x_0^B,y)
  \label{eq:matched-reward-gap}
\end{equation}
provides a local preference without claiming global superiority of either
route. For continuous rewards (PickScore), ties are rare; an
\emph{adaptive flipping} guard (Figure~\ref{fig:framework}c) requires at
least half of the pull targets to come from route $A$, flipping the
lowest-margin route-$B$ win when the ratio falls below $0.5$. This guard
keeps the coverage-restoring $b\theta\theta$ branch dominant and prevents
the hacking-prone $\theta\theta u$ branch from re-hacking the pull
distribution. For rule-based discrete rewards (GenEval), ties are frequent;
we break them with a near 50/50 split that slightly favors route $A$, without
activating the flipping guard.

\subsection{Reward-Ranked Joint-and-Paired Repair}
\label{sec:paired-forward}
We realize the matched preference with the NFT mirror mechanism from
Section~\ref{sec:prelim-nft}. The loss in v-space is
\begin{equation}
  \mathcal{L}_{\mathrm{repair}}=
  \underbrace{\left\|v_\theta^{-}(x_t^L,t,y)-v^L\right\|_2^2}_{\text{mirrored push}}
  +\underbrace{\left\|v_\theta^{+}(x_t^H,t,y)-v^H\right\|_2^2}_{\text{ordinary pull}}.
  \label{eq:repair-loss}
\end{equation}
The mirror reference $v_\theta^{-}$ uses an EMA-smoothed copy of the policy
with constant decay, since repair starts from a late-stage checkpoint.

For each paired timestep, we sample a fresh perturbation $\xi$ and construct
\begin{equation}
  x_t^H=(1-\sigma_t)x_0^H+\sigma_t\xi, \quad
  x_t^L=(1-\sigma_t)x_0^L+\sigma_t\xi.
  \label{eq:paired-renoise}
\end{equation}
with targets $v^H=\xi-x_0^H$ and $v^L=\xi-x_0^L$. The pull and push share
$\xi$, treating corruption as a matched nuisance; sampling $\xi$ independently of $\eps$ avoids a per-(prompt, noise) fitting
shortcut~\cite{wen2024detecting}.

At $\sigma_t=1$, a fresh $\xi$ would break the coupling, so we use the
native rollout noise $\eps$ for a \emph{joint} update
(Figure~\ref{fig:framework}c), with $x_t^H=x_t^L=\eps$ and targets
$v^H=\eps-x_0^H$, $v^L=\eps-x_0^L$. This anchor matters because early
denoising decisions strongly influence global
structure~\cite{jin2026stagewise}. Each repair step applies one joint update
with $\eps$ at $\sigma_t=1$, followed by an even number of paired updates
with fresh $\xi$ at $\sigma_t<1$ (see the appendix for ablations on the
joint-noise choice and rollout length). Mixed routes only construct training
supervision; inference uses the standard conditional forward pass with no
routing overhead.

\section{Experiments}
\label{sec:exp}

We evaluate \textbf{ReNFT} on PickScore and GenEval: standard reward
post-training produces the collapsed checkpoint that motivates repair
(Section~\ref{sec:exp-quantitative}), \textbf{ReNFT} recovers diversity while
retaining reward both quantitatively and qualitatively
(Sections~\ref{sec:exp-quantitative} and~\ref{sec:exp-qualitative}), and
controlled ablations identify the responsible design decisions
(Section~\ref{sec:exp-ablation}).

\begin{table*}[!t]
  \centering
  \renewcommand{\arraystretch}{1.08}
  \resizebox{0.8\textwidth}{!}{%
  \begin{tabular}{lccccccc}
    \toprule
    \multirow{2}{*}{\textbf{Method}} &
    \multicolumn{4}{c}{\textbf{Model-Based Reward}} &
    \multicolumn{3}{c}{\textbf{Diversity}} \\
    \cmidrule(lr){2-5} \cmidrule(lr){6-8}
    & PickScore $\uparrow$ & Aesthetic $\uparrow$ & ImageReward $\uparrow$ &
    HPSv2 $\uparrow$ & LPIPS-Div $\uparrow$ & DreamSim-Div $\uparrow$ &
    DINO-Div $\uparrow$ \\
    \midrule
    SD3.5-M & \gref{21.73} & \gref{6.026} & \gref{1.06} & \gref{0.295} &
      \gref{0.599} & \gref{0.242} & \gref{0.251} \\
    \midrule
    + Flow-GRPO & \cellcolor{graybg}22.90 & 6.277 & 1.31 & 0.305 &
      0.463 & 0.122 & 0.134 \\
    + DiffusionNFT & \cellcolor{graybg}\textbf{23.51} & \textbf{6.592} &
      \textbf{1.44} & \underline{0.327} & 0.430 & 0.119 & 0.112 \\
    + DiverseGRPO & \cellcolor{graybg}22.99 & 6.234 & 1.27 &
      \textbf{0.332} & \underline{0.506} & \underline{0.163} & \underline{0.145} \\
    + E2PO$^\dagger$ & \cellcolor{graybg}\underline{23.38} & \underline{6.538} & 1.29 &
      0.325 & -- & -- & -- \\
    + \textbf{Ours} & \cellcolor{graybg}23.26 &
      6.344 & \underline{1.41} & 0.323 & \textbf{0.565} &
      \textbf{0.189} & \textbf{0.182} \\
    \bottomrule
  \end{tabular}%
  }
  \caption{\textbf{Quantitative results on the PickScore test set.}
  \colorbox{graybg}{Shaded}: in-domain training reward.
  Parenthesized row: reference only, excluded from ranking.
  \textbf{Bold}: best; \underline{underline}: second best.
  $^\dagger$See Section~\ref{sec:exp-setup}.}
  \label{tab:main-pickscore}
\end{table*}

\begin{table}[!tb]
  \centering
  \renewcommand{\arraystretch}{1.08}
  \resizebox{\columnwidth}{!}{%
  \begin{tabular}{lcccc}
    \toprule
    \multirow{2}{*}{\textbf{Method}} &
    \multicolumn{1}{c}{\textbf{Rule-Based Reward}} &
    \multicolumn{3}{c}{\textbf{Diversity}} \\
    \cmidrule(lr){2-2} \cmidrule(lr){3-5}
    & GenEval $\uparrow$ & LPIPS-Div $\uparrow$ & DreamSim-Div $\uparrow$ &
    DINO-Div $\uparrow$ \\
    \midrule
    SD3.5-M & \gref{0.654} & \gref{0.687} & \gref{0.339} & \gref{0.379} \\
    \midrule
    + Flow-GRPO & \cellcolor{graybg}0.802 & 0.343 & 0.176 & 0.205 \\
    + DiffusionNFT & \cellcolor{graybg}\textbf{0.946} & 0.292 & 0.149 & 0.186 \\
    + DiverseGRPO & \cellcolor{graybg}0.869 & \underline{0.454} & \underline{0.198} & \underline{0.246} \\
    + E2PO$^\dagger$ & \cellcolor{graybg}0.932 & -- & -- & -- \\
    + \textbf{Ours} &
      \cellcolor{graybg}\underline{0.937} & \textbf{0.496} &
      \textbf{0.231} & \textbf{0.295} \\
    \bottomrule
  \end{tabular}%
  }
  \caption{\textbf{Quantitative results on the GenEval test set.}
  \colorbox{graybg}{Shaded}: in-domain training reward.
  Parenthesized row: reference only, excluded from ranking.
  \textbf{Bold}: best; \underline{underline}: second best.
  $^\dagger$See Section~\ref{sec:exp-setup}.}
  \label{tab:main-geneval}
\end{table}

\subsection{Experimental Setup}
\label{sec:exp-setup}
We use SD3.5-M~\cite{sd3} with LoRA adapters and train separate
PickScore~\cite{pickscore} and GenEval~\cite{geneval} checkpoints.
\textbf{ReNFT} reloads the DiffusionNFT LoRA at step 700 (PickScore) or 250
(GenEval) and repairs for 50 steps with the default configuration in the
appendix. Model-based rewards use all 1,024 PickScore test prompts; the
rule-based reward uses all 553 GenEval prompts; diversity uses 100 fixed
prompts with 12 fixed-noise samples each. NFT, E$^2$PO, and \textbf{ReNFT}
use CFG=1; base and GRPO-style checkpoints use CFG=4.5, so cross-CFG
comparisons are end-to-end method comparisons, not controlled CFG ablations.

We compare the untrained SD3.5-M reference, Flow-GRPO~\cite{liu2025flowgrpo},
DiffusionNFT (NFT)~\cite{zheng2025diffusionnft},
DiverseGRPO~\cite{liu2025diversegrpo}, and E$^2$PO~\cite{hu2026e2po}.
E$^2$PO is not open-sourced; we match its official training budget and reuse
its paper-reported reward values (marked $^\dagger$), omitting it from the
diversity ranking. Quality metrics are PickScore-v1,
Aesthetic~\cite{radford2021learning,schuhmann2022laion},
ImageReward-v1.0~\cite{imagereward2023}, and HPSv2.1~\cite{hpsv2}. Diversity
metrics are LPIPS-Div~\cite{lpips}, DreamSim-Div~\cite{fu2023dreamsim}, and
DINOv3-Div~\cite{simeoni2025dinov3}, each computed as the mean pairwise
distance among 12 samples
per prompt. Anti-hub uses 24 probes and 1,000 candidates; further details
are in the appendix.

\subsection{Quantitative Results}
\label{sec:exp-quantitative}
\noindent\textbf{Reward post-training produces the repair setting.}
Tables~\ref{tab:main-pickscore} and~\ref{tab:main-geneval} confirm the
reward--diversity tension hypothesized in Section~\ref{sec:repair-view}: the
untrained base (shown in parentheses as a
reference row, excluded from ranking) has the largest within-prompt
diversity but the weakest task reward, while the continued
DiffusionNFT run reaches the highest PickScore and GenEval values at steps
750 and 300, with DreamSim-Div falling from 0.242 to 0.119 and from
0.339 to 0.149, respectively. \textbf{ReNFT} instead branches from the earlier NFT
checkpoints $H$ at steps 700 and 250 under the same remaining 50-step budget.
Among the compared post-trained methods, \textbf{ReNFT} attains the highest
diversity in all three metrics under both protocols.


\noindent\textbf{ReNFT recovers diversity while retaining reward.}
Relative to NFT at the protocol endpoint, \textbf{ReNFT} retains 98.9\% of
PickScore (23.26 vs. 23.51) and 99.0\% of GenEval (0.937 vs. 0.946), while
improving DreamSim-Div by 58.8\% (0.189 vs. 0.119) and 55.0\% (0.231 vs.
0.149), respectively. The trade-off is asymmetric: roughly 1\% reward cost
versus 55--59\% diversity gain. \textbf{ReNFT} ranks second on GenEval reward
and PickScore ImageReward; all three diversity metrics are best among
compared post-trained methods under both protocols. This asymmetry reflects
how susceptible each reward is to hacking: model-based scores of PickScore
depend
heavily on exploiting the model preference, so an anti-hacking method faces
a harder reward landscape, whereas GenEval's rule-based scoring rewards
correct color, position, and count through segmentation, leaving less room
for hacking-specific artifacts.

\noindent\textbf{Repair dynamics.} In
Figure~\ref{fig:training-dynamics}(a)--(d), under PickScore reward decreases
with local oscillations toward the table value, while diversity rises
rapidly, peaks around step 35, and decays mildly, remaining above every other
post-trained baseline at step 50. The rise--peak--decay pattern reflects the
dual-route win ratio: $b\theta\theta$ initially dominates $\theta\theta u$,
but as the routes converge the ratio oscillates near $0.5$; adaptive flipping
maintains a minimum $b\theta\theta$ pull share, yet the unconditional branch
gradually re-hacks, causing diversity to recede and reward to climb again
along a hacking trajectory. GenEval shows no turning point within 50 steps:
its
wider acceptable reward range produces more stable curves: reward drops
less, diversity rises more, and the dual-branch balance point is reached
later.

\noindent\textbf{Comparison with diversity-oriented baselines.}
DiverseGRPO preserves more diversity than NFT but at a larger reward cost:
97.8\% of PickScore (22.99 vs. 23.51) and 91.9\% of GenEval (0.869 vs.
0.946), compared to \textbf{ReNFT}'s 98.9\% and 99.0\%. E$^2$PO's reused
reward enters the ranking with provenance marked $^\dagger$; its
interface-level bias correction is complementary to our internal-route
approach.

\subsection{Qualitative Results}
\label{sec:exp-qualitative}
Figures~\ref{fig:qual-geneval} and~\ref{fig:qual-pickscore} compare Base,
NFT, DiverseGRPO, and \textbf{ReNFT} on GenEval+OCR and PickScore prompts,
respectively, with a shared noise set.

\noindent\textbf{PickScore prompts.} NFT and DiverseGRPO exhibit recognizable
hacking tendencies: a recurring color tone, densely packed elements, and a
preference for single female subjects. NFT's late-stage outputs respond
primarily to prompt changes, becoming insensitive to the initial noise.
\textbf{ReNFT} restores scene- and style-level variation across seeds without
reverting to the base: dog statues vary in pose and setting, wedding scenes
recover diverse compositions, and the emperor-and-robots prompt produces
distinct arrangements.

\noindent\textbf{GenEval and OCR prompts.} NFT collapses counting-related
prompts onto pure white backgrounds, a hacking shortcut that simplifies
segmentation and counting. \textbf{ReNFT} removes this bias: count, position,
and color remain correct while backgrounds and styles vary. The OCR prompts
are out-of-distribution for our PickScore-trained repair: NFT and DiverseGRPO
retain some text-rendering ability from OCR-adjacent prompts, but hacking
distorts style and layout. \textbf{ReNFT} produces more legible and varied
text, approaching base-level diversity with higher text fidelity. These
observations confirm the repair-view hypothesis
(Section~\ref{sec:repair-view}): suppressed modes were compressed, not
deleted. A diagnostic view from unconditional (CFG=0) samples is provided in
the appendix.

\subsection{Ablation Study}
\label{sec:exp-ablation}
The ablations examine two design decisions in the main text: anti-hub prompt
prioritization and internal route construction. All variants branch from the
same checkpoint $H$ under the same 50-step budget and checkpoint-evaluation
protocol.

\noindent\textbf{Anti-hub prompt prioritization.}
The controlled variant replaces anti-hub prompts with random prompts
from the same candidate pool, preserving the prompt count, sampling
budget, and update rule. The two variants track nearly
identical reward trajectories, whereas diversity diverges sharply: without
anti-hub selection, diversity rises only briefly and then decays back below its
starting level, while the full configuration sustains the recovery. The
PickScore training distribution is itself biased toward the hacking
preferences of the reward model, so $\theta\theta u$ may not be substantially worse than
$b\theta\theta$ from the start; without anti-hub prompts that expose the
bias most distinguishably, there is little diversity-recovery momentum.

\noindent\textbf{Internal route construction.}
Figure~\ref{fig:training-dynamics}(e,f) compares
$(b\theta,\theta u)$, $(b\theta\theta,\theta\theta u)$, and longer periodic
variants under the same budget. With pattern length 2, the base and
unconditional content is too high: $\theta u$ consistently produces
low-reward outputs, the win ratio stays near $1.0$, and saturation/contrast
anomalies cause reward to collapse while diversity rises from degenerate
outputs. Longer blocks approach the NFT direction. The three-step pattern
$(b\theta\theta,\theta\theta u)$ achieves the clearest diversity recovery
with the least reward sacrifice.

Appendix ablations support the default rollout length, joint-noise choice,
and EMA decay; sensitivity to the anti-hub pool size and flipping threshold
remains future work.

\section{Conclusion}
We studied reward-induced mode collapse as internal probability-mass
reallocation, where post-training concentrates mass on reward-favored modes
and suppressed alternatives remain accessible through internal routes of the
generator. \textbf{ReNFT} operationalizes this view by prioritizing
anti-hub prompts with unconditional probes, generating matched
counterfactuals from the same prompt and noise through two policy-dominated
mixed routes, and assigning pull and push roles through reward ranking with
an adaptive flipping guard, all realized through joint-and-paired NFT
updates. Across PickScore and GenEval, \textbf{ReNFT} retains 98.9\% and
99.0\% of NFT's reward while improving DreamSim-Div by 58.8\% and 55.0\%,
confirming that collapsed adapters can be repaired from within without
external diversity objectives, text-encoder modification, or sacrificing the
acquired reward. This internal repair perspective opens a complementary
direction to external interventions, extending naturally to other
post-training paradigms and backbones.

\clearpage
\bibliography{refs}

\clearpage
\appendix
\section{Experimental Details}
\label{sec:exp-details}
\noindent\textbf{Training configuration.} SD3.5-M~\cite{sd3} runs train LoRA adapters
(rank 32, alpha 64) with learning rate $10^{-4}$, 48 unique samples per
epoch, and group size 24. All runs use seed
42, the same seed as prior work~\cite{hu2026e2po}. Repair uses off-policy
training with a constant EMA decay of $0.25$, $N_i{=}36$ rollout steps, and
$N_t{=}7$ training timesteps per update (one native-noise joint anchor and six
fresh-noise paired updates). The minimum route-$A$ pull ratio is $0.5$.

\noindent\textbf{Prompt splits and tuning protocol.}
PickScore~\cite{pickscore} provides 15{,}471 training prompts and
GenEval~\cite{geneval} provides 33{,}199; evaluation uses the full held-out
test sets as in the main paper. Anti-hub candidate pools are drawn from the
training bank, and all hyperparameter selection reported in
Section~\ref{sec:ablations} was based on evaluation trajectories of saved
checkpoints on the training prompts (not training-metric logs), with the
held-out prompts never used for tuning.

\noindent\textbf{Training budget and checkpoint choice.} The total budgets
(750 steps for PickScore, 300 for GenEval) follow the official configuration
of E$^2$PO~\cite{hu2026e2po}. All trained baselines share these budgets and
are trained to the same endpoints. \textbf{ReNFT} branches from the NFT
checkpoint 50 steps before the endpoint and uses the remaining 50 steps for
repair, matching the continued-training baselines step for step. The 50-step
budget is motivated by the pull-ratio analysis in
Section~\ref{sec:win-ratio}.

\noindent\textbf{Evaluation settings.} On SD3.5-M, evaluation uses
512$\times$512 resolution with 40-step ODE sampling. NFT-family checkpoints
(NFT, E$^2$PO, and \textbf{ReNFT}) are evaluated at CFG$=$1 (i.e., CFG-free),
following the protocol of DiffusionNFT~\cite{zheng2025diffusionnft}.

\noindent\textbf{Diversity metrics.} For each held-out prompt, we generate 12
samples from fixed initial noises. LPIPS~\cite{lpips} uses the AlexNet backbone
with bicubic resizing to 256$\times$256; DreamSim~\cite{fu2023dreamsim} uses
the official pretrained model and cosine distance between its embeddings;
DINOv3~\cite{simeoni2025dinov3} uses the ViT-B/16 LVD-1689M CLS embedding with
Resize(256)--CenterCrop(224) preprocessing and cosine distance. For every
metric, we average all $\binom{12}{2}=66$ pairwise distances within a prompt,
then average the resulting prompt-level scores over the 100 held-out prompts.

\noindent\textbf{Quality metrics.} PickScore~\cite{pickscore},
Aesthetic~\cite{radford2021learning,schuhmann2022laion},
ImageReward~\cite{imagereward2023}, and HPSv2~\cite{hpsv2} are computed as
in the main paper.

\noindent\textbf{Algorithm summary.} Algorithm~\ref{alg:renft} summarizes the
repair loop; all symbols follow Sections 4.2--4.4 of the main paper, and
concrete hyperparameter values are those given above.

\begin{algorithm}[!tb]
   \caption{Repair NFT (\textbf{ReNFT}).}
   \label{alg:renft}
   \small
   \textbf{Require:} Hacked checkpoint $\theta_H$, frozen base route $b$,
   reward model $R(\cdot)$, frozen SigLIP2 encoders $e_{\mathrm{text}},
   e_{\mathrm{img}}$, training prompt bank $\mathcal{Y}$, routing patterns
   $P_A, P_B$ with rollout operator $\mathcal{T}$, probe count $M$, candidate
   number $C$, training timesteps $N_t$, learning rate $\lambda$, EMA decay
   $\eta$, minimum route-$A$ pull ratio $\tau$.\\
   \textbf{Initialize:} $\theta\gets\theta_H$;
   $\theta^{\mathrm{old}}\gets\theta_H$; batch buffer $\mathcal{B}\gets\varnothing$.
   \begin{algorithmic}[1]
    \FOR{each optimizer step}
        \STATE Sample $M$ unconditional probes
        $\{u_j\}_{j=1}^{M}$
        \Comment{anti-hub selection}
        \STATE Score $C$ candidate prompts by
        $s(y)\gets\frac{1}{M}\sum_{j=1}^{M}
        \langle e_{\mathrm{text}}(y),e_{\mathrm{img}}(u_j)\rangle$
        \STATE Select anti-hub set:
        $\mathcal{Y}_{\mathrm{ah}}\gets
        \operatorname*{Bottom}_{C,\,y\in\mathcal{Y}}s(y)$ \hfill Eq.~(6)
        \FOR{$y\in\mathcal{Y}_{\mathrm{ah}}$}
            \STATE Draw $\varepsilon\sim\mathcal{N}(0,I)$
            \Comment{matched rollouts}
            \STATE Compute matched endpoints
            $x_0^A\gets\mathcal{T}_{P_A,y}(\varepsilon)$,
            $x_0^B\gets\mathcal{T}_{P_B,y}(\varepsilon)$
            \STATE Rank by reward:
            $H_y\gets\arg\max_{k\in\{A,B\}}R(x_0^k,y)$;
            $L_y\gets\arg\min_{k\in\{A,B\}}R(x_0^k,y)$ \hfill Eq.~(8)
            \STATE Collect
            $(y,\varepsilon,x_0^{H_y},x_0^{L_y})$ into $\mathcal{B}$
        \ENDFOR
        \STATE $\mathcal{B}\gets\operatorname{GuardFlip}(\mathcal{B},\tau)$
        \Comment{minimum route-$A$ pull share}
        \FOR{$(y,\varepsilon,x_0^H,x_0^L)\in\mathcal{B}$}
            \FOR{$t\in\{1,\ldots,N_t\}$}
                \STATE $\xi\gets\varepsilon$ if $t{=}1$, else
                $\xi\sim\mathcal{N}(0,I)$
                \Comment{self / fresh noise}
                \STATE Re-noise high endpoint:
                $x_t^H\gets(1{-}\sigma_t)x_0^H{+}\sigma_t\xi$
                \STATE Re-noise low endpoint:
                $x_t^L\gets(1{-}\sigma_t)x_0^L{+}\sigma_t\xi$
                \STATE Targets:
                $v^H\gets\xi{-}x_0^H$;
                $v^L\gets\xi{-}x_0^L$ \hfill Eq.~(10)
                \STATE Implicit positive velocity:
                $v_\theta^{+}\gets v_\theta(x_t^H,t,y)$ \hfill Eq.~(2)
                \STATE Implicit negative velocity:
                $v_\theta^{-}\gets 2\,v_{\mathrm{old}}(x_t^L,t,y)
                -v_\theta(x_t^L,t,y)$ \hfill Eq.~(3)
                \STATE Positive loss:
                $\mathcal{L}^{+}\gets
                \|v_\theta^{+}-v^H\|_2^2$
                \STATE Negative loss:
                $\mathcal{L}^{-}\gets
                \|v_\theta^{-}-v^L\|_2^2$
                \STATE $\theta\gets\theta-
                \lambda\nabla_\theta(\mathcal{L}^{+}{+}\mathcal{L}^{-})$
                \hfill Eq.~(9)
            \ENDFOR
        \ENDFOR
        \STATE $\theta^{\mathrm{old}}\gets
        \eta\,\theta^{\mathrm{old}}+(1{-}\eta)\,\theta$
        \Comment{EMA update}
    \ENDFOR
   \end{algorithmic}
   \textbf{Output:} $v_\theta$
\end{algorithm}


\section{Additional Ablations}
All ablations in this section use the SD3.5-M backbone under the PickScore
protocol, branching from the same hacked checkpoint $H$ as the main
experiments.
\label{sec:ablations}

\begin{figure}[!t]
  \centering
  \includegraphics[width=0.9\linewidth]{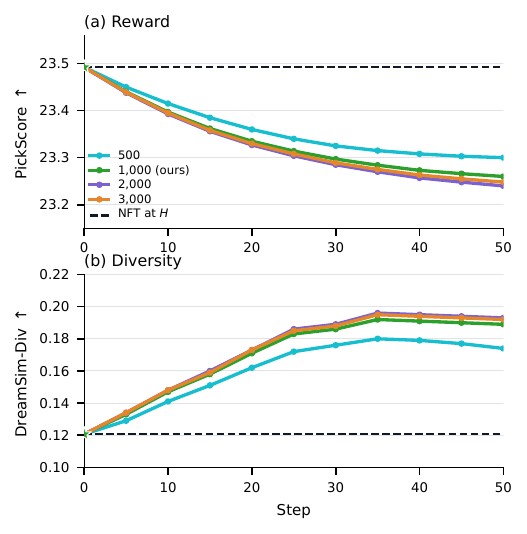}
  \caption{\textbf{Ablation of Anti-Hub Candidate Pool Size on SD3.5-M (PickScore).} The
  green trajectory (1,000 candidates) is identical to the default
  configuration in all other ablation figures.}
  \label{fig:antihub-pool}
\end{figure}
\begin{figure}[!tb]
  \centering
  \includegraphics[width=0.9\linewidth]{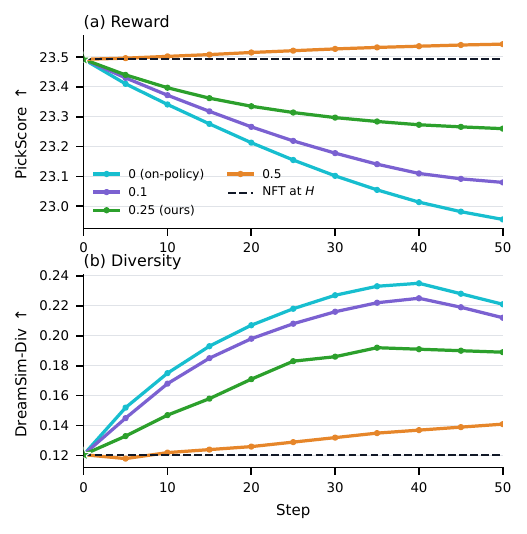}
  \caption{\textbf{Ablation of EMA Decay on SD3.5-M (PickScore).} Larger decays preserve
  reward but barely recover diversity; smaller decays recover more diversity
  at larger reward cost, with on-policy training (decay $0$) reaching the
  highest diversity but the lowest reward. The $0.25$ curve is Ours.}
  \label{fig:ema-decay}
\end{figure}
\begin{figure}[!tb]
  \centering
  \includegraphics[width=0.9\linewidth]{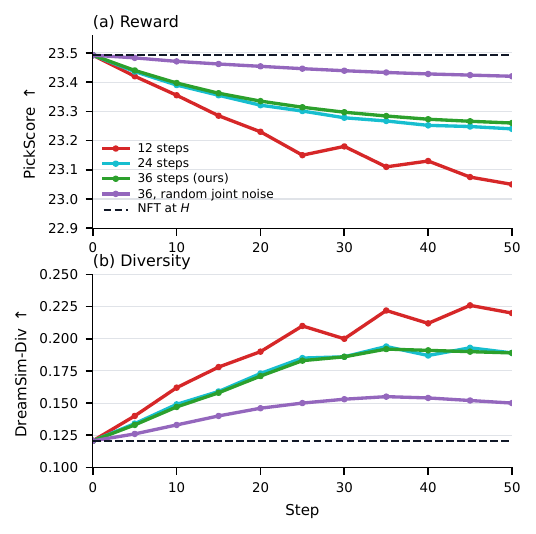}
  \caption{\textbf{Ablation of Rollout Length and Joint-Step Self-Noise on SD3.5-M (PickScore).}
  The 36-step curve is Ours; all variants branch from the same checkpoint $H$.}
  \label{fig:sampling-steps}
\end{figure}

\subsection{Ablation of Anti-Hub Selection and Pool Size}
\label{sec:anti-hub-sensitivity}
Anti-hub selection uses 24 unconditional probe generations to score each of
1,000 candidate prompts with SigLIP2~\cite{tschannen2025siglip2} text--image
embeddings, selecting the lowest-mean-similarity prompts for repair. The on/off ablation in the main paper confirms that replacing anti-hub-selected prompts with
random ones collapses diversity recovery while reward tracks closely,
isolating prompt prioritization from prompt coverage.

\paragraph{Candidate pool size.}
Figure~\ref{fig:antihub-pool} varies the candidate pool size
$\{500, 1000, 2000, 3000\}$ under the same 50-step repair budget. Pools of
1,000 or more candidates yield nearly identical reward and diversity
trajectories, indicating that coverage saturates quickly; the smallest pool
(500) retains reward best but recovers noticeably less diversity due to
insufficient prompt coverage. The default 1,000 candidates are sufficient.

\subsection{Ablation of EMA Decay}
\label{sec:ema-sensitivity}

We compare constant decays $\{0, 0.1, 0.25, 0.5\}$ under the same hacked
checkpoint, 50-step repair budget, and checkpoint-evaluation protocol, where
$0$ corresponds to on-policy training without the off-policy reference. Large
decay ($0.5$) slows the reference so much that diversity barely improves while
reward instead rises slightly, reproducing an NFT-like hacking direction.
Smaller decays move along the trade-off in the opposite direction: decay
$0.1$ recovers substantially more diversity at a visible reward cost, and
fully on-policy training (decay $0$) reaches the highest diversity overall
but the lowest reward, showing that without an off-policy reference the
mirrored push drifts too far from the current policy.
Decay $0.25$ gives the best
reward--diversity trade-off and is used in all main experiments.

\subsection{Ablation of Rollout Length and Joint-Step Self-Noise}
\label{sec:sampling-sensitivity}

\noindent\textbf{Rollout length.}
Figure~\ref{fig:sampling-steps} varies the rollout length. With 12 steps,
reward drops quickly but diversity rises the highest, indicating that a
short, coarse rollout leaves samples under-converged toward the high-reward
mode. The 24- and 36-step variants are close on both metrics, with the
36-step rollout ending marginally higher on reward; since repair needs only
about 50 optimizer steps, the extra sampling overhead is modest. We use 36
steps in the main configuration.

\noindent\textbf{Self-noise at the joint step.}
At the joint (first) training timestep, ReNFT reuses the initial noise of the
rollout $\varepsilon$ rather than drawing fresh noise. Replacing this self-noise
with fresh random noise at the joint step keeps reward higher but markedly
weakens diversity recovery, consistent with aggravated hacking: fresh noise
decouples the pull and push endpoints from the shared counterfactual
trajectory. We retain self-noise at $\sigma_t{=}1$. Loading self-sampled noise
at every training timestep was unstable and diverged, so we exclude that
variant from the controlled comparison.

\subsection{Ablation of Route Position and Contiguity}
\label{sec:route-position}
The main paper uses fixed route patterns in which the base step $b$ occupies
the first position of each block and the unconditional step $u$ occupies the
last. This section explains the design rationale.

\paragraph{Why $b$ at the block start.} Early denoising steps have the
strongest influence on global structure~\cite{jin2026stagewise}: the vector
field at high noise determines layout and composition before local details.
Placing $b$ at the block start injects a base-like direction where it can
most effectively steer toward suppressed structural alternatives. A base step
at the block end, by contrast, would leave no subsequent $\theta$ step to
correct base-quality artifacts (blur, incomplete objects), which the reward
ranking should filter out rather than recover.

\paragraph{Why $u$ at the block end.} The unconditional route exposes
prompt-independent bias (recurring textures, palettes, stylistic artifacts).
Placing $u$ at the block end lets it act on a nearly-formed image where this
bias is observable. An earlier $u$ would steer toward the unconditional
hacking mode before the conditional route establishes prompt-specific
structure, yielding a pull ratio locked near $1.0$; the pattern-length-2
ablation in the main paper shows this leads to degenerate outputs.

\paragraph{Why no consecutive $bb$ or $uu$ blocks.} The core principle is
that $b$ and $u$ serve to inject diversity or expose bias, but they must not
dominate the trajectory. Consecutive base steps ($bb$) push the sampling
manifold toward the base distribution for an extended interval, effectively
reverting to base-like sampling and losing the acquired reward. Consecutive
unconditional steps ($uu$) amplify the prompt-independent bias for an
extended interval, strengthening the very hacking direction we aim to
suppress. In both cases the trajectory deviates too far from the current
policy, destabilizing training and degrading both reward and diversity.

\paragraph{Toward a minimal repeating pattern.} The above considerations
converge on a simple design: place $b$ at the start and $u$ at the end,
avoid consecutive $bb$ or $uu$ blocks, and keep the majority of steps on the
conditional route $\theta$. The most decoupled realization is to define the
shortest pattern that satisfies these constraints and repeat it uniformly
across the inference timesteps, which is exactly what the default
$(b\theta\theta,\theta\theta u)$ configuration does. A systematic ablation
over alternative positions (e.g., $\theta\theta b$, $\theta u\theta$) and
consecutive blocks (e.g., $bb\theta\theta$) would further validate these
choices and is left to future work.

\section{Repair Dynamics and Diagnostics}
\label{sec:dynamics}

\subsection{Dual-Route Pull-Ratio Dynamics}
\label{sec:win-ratio}

\begin{figure*}[!t]
  \centering
  \includegraphics[width=0.8\textwidth]{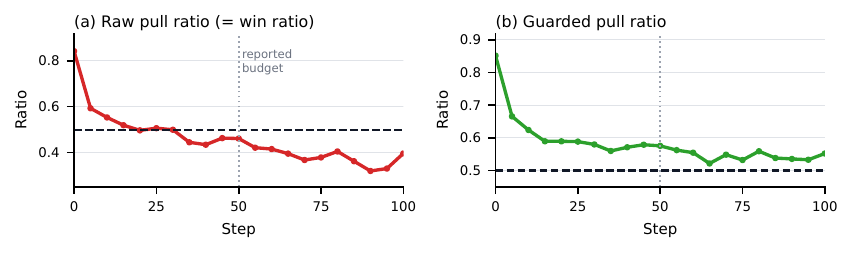}
  \caption{\textbf{Dual-Route Pull Ratios during SD3.5-M PickScore Repair.} (a) The
  raw route-$A$ ($b\theta\theta$, base-probing) pull ratio, which equals route
  $A$'s win rate in the per-pair reward ranking before the adaptive flipping
  guard; (b) the guarded ratio after the guard enforces the $\geq 0.5$
  per-group floor. The vertical dotted line marks the reported 50-step
  budget.}
  \label{fig:win-ratio}
\end{figure*}

Figure~\ref{fig:win-ratio} reports the dual-route pull ratios over 100 repair
steps, extending beyond the reported 50-step budget. The raw route-$A$ pull
ratio (route $A$'s win rate before any flipping) starts at $0.84$, drops
below $0.5$ around step $20$, and settles in the $0.32$--$0.42$ band from
step $50$ onward, ending at $0.40$. This trajectory matches the diversity peak
near step $35$ (Figure~2 of the main paper): as repair progresses, the
unconditional route increasingly wins the ranking, i.e., it gradually
re-hacks the reward model. The guarded ratio holds within $0.52$--$0.62$
from step $10$ onward and never breaches the $0.5$ floor, with the gap to the
raw curve widening as the guard flips more low-margin pairs. The same
pull-mass reasoning beyond step $50$ (continued repair draws pull mass from
guard-flipped, lower-margin pairs while diversity has already passed its
peak) directly motivates terminating repair at $50$ steps.

\subsection{Unconditional Sampling Analysis}
\label{sec:uncond-display}

\begin{figure*}[!t]
  \centering
  \includegraphics[width=0.9\textwidth]{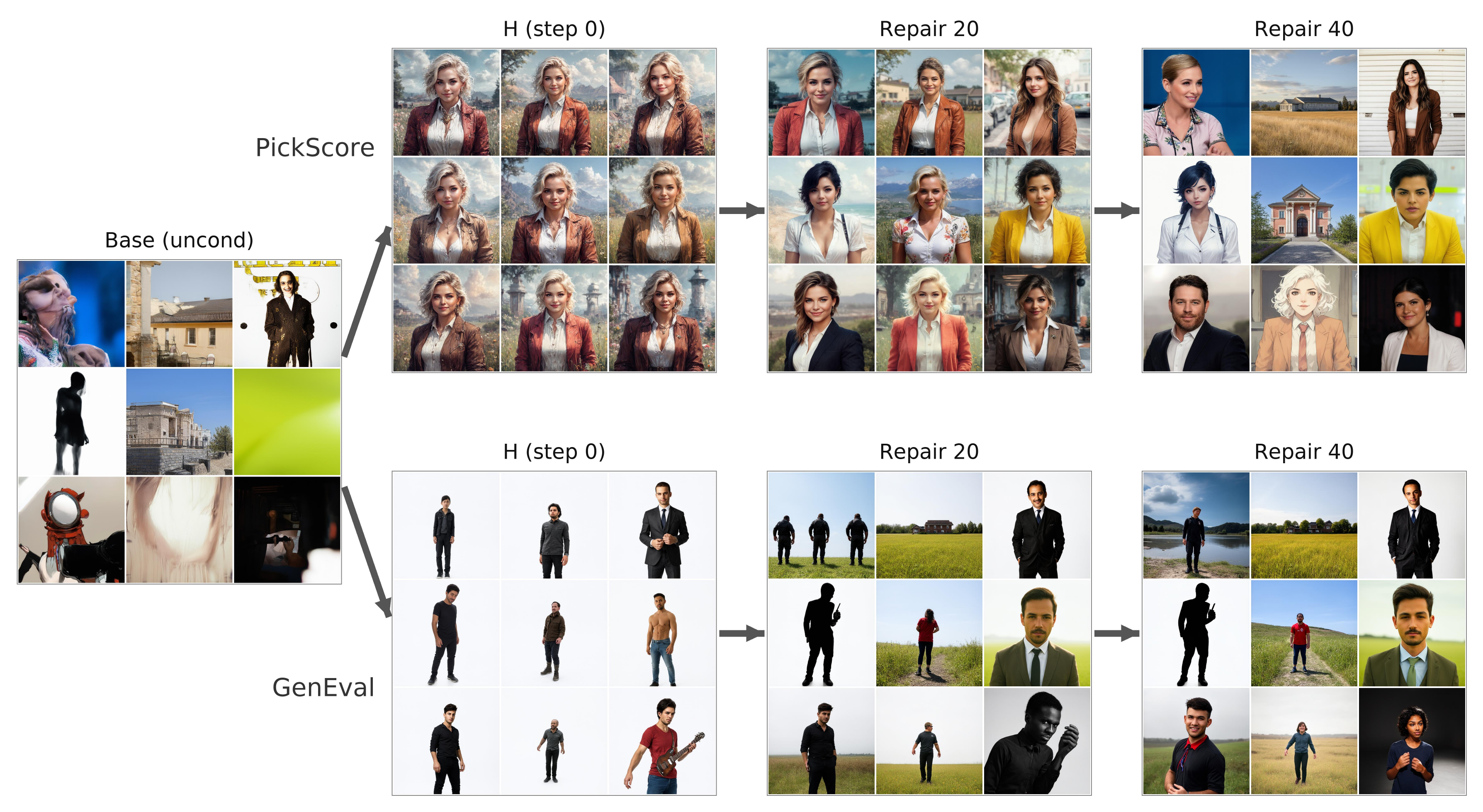}
  \caption{\textbf{Unconditional Sample Display across NFT Training and ReNFT Repair on SD3.5-M.}
  Empty-prompt (CFG=0) samples from a fixed noise grid. The base generator
  branches into PickScore (top) and GenEval (bottom) repair trajectories, shown
  at the hacked checkpoint $H$ and repair steps 20 and 40.}
  \label{fig:uncond-display}
\end{figure*}

Figure~\ref{fig:uncond-display} tracks the same fixed noise indices across
training and repair on SD3.5-M. The base generator spans unrelated subjects,
scenes, styles, and palettes (Figure~1 of the main paper), whereas checkpoint
$H$ maps the grid to protocol-specific hubs. Under PickScore, the nine samples
concentrate on warm-toned, highly detailed female portraits with similar
framing and backgrounds; under GenEval, they concentrate on isolated,
full-body figures against plain backgrounds. Thus the two rewards induce
different visual signatures, but both sharply reduce unconditional diversity.
Repair progressively reopens the grid: step $20$ broadens appearance and
context, and step $40$ produces more varied subjects, compositions,
backgrounds, and palettes under the same noise indices. The repaired grids
remain between the narrow $H$ hubs and the wider base range rather than simply
reverting to Base. Because the same noise indices are shared across stages,
each repaired sample partially retains the layout, style, and palette of the
corresponding base sample (e.g., a blue background in Base tends to stay
blue through $H$ and repair), and the correspondence strengthens as diversity
recovers: the more varied grids at step $40$ align more closely with Base
than the collapsed grids at $H$. This same-noise contraction-and-reopening
pattern supports
the hub-reading use of the unconditional route (Section~4.2 of the main paper)
and the suppression-not-deletion hypothesis (Section~4.1 of the main paper).

\begin{figure*}[!t]
  \centering
  \includegraphics[width=0.9\textwidth]{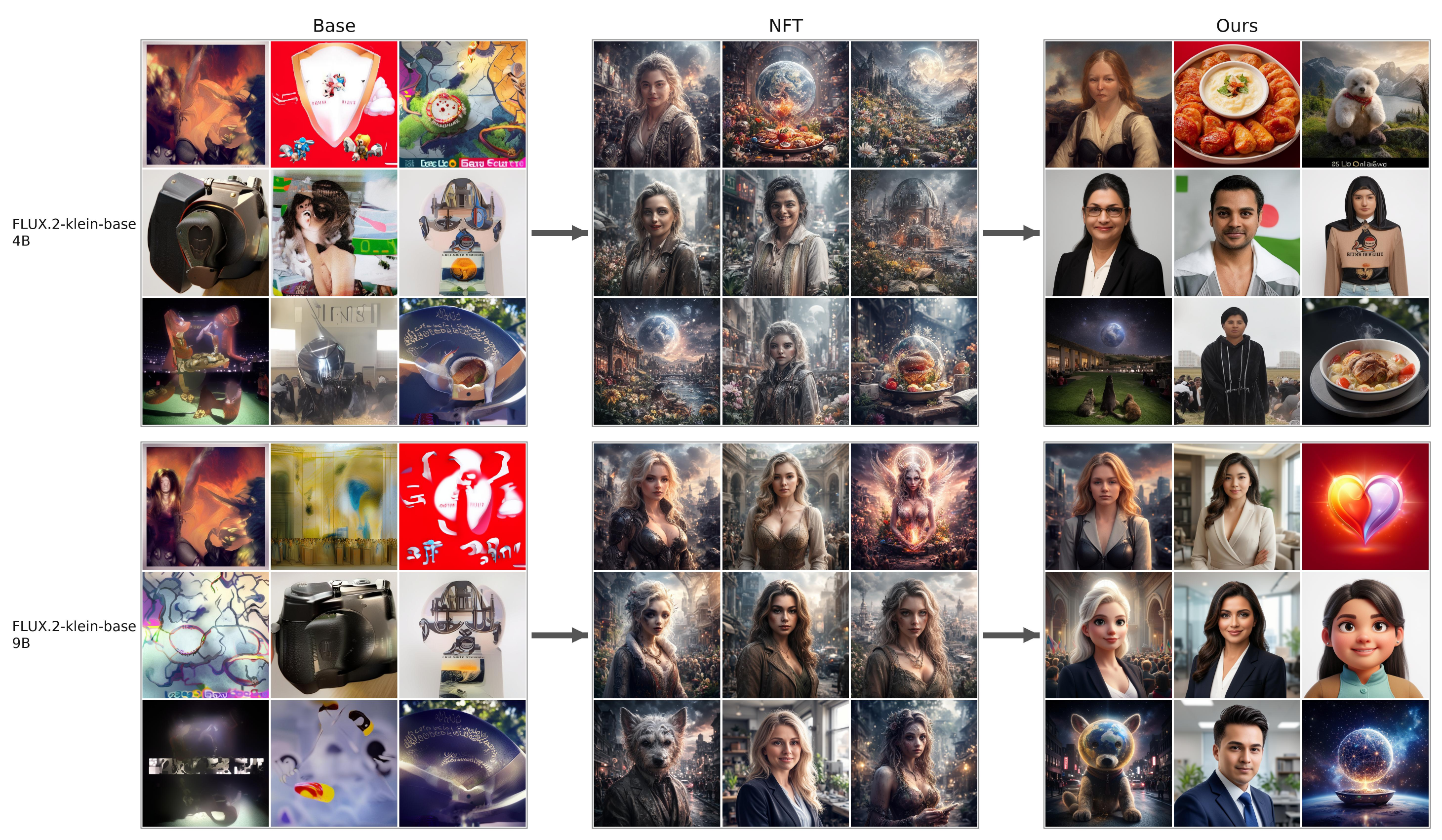}
  \caption{\textbf{Unconditional Sample Display on FLUX.2-klein-base.} Empty-prompt samples from a fixed noise grid for the base model, the NFT checkpoint, and \textbf{ReNFT} (4B backbone on top, 9B at the bottom).}
  \label{fig:flux-uncond}
\end{figure*}

Figure~\ref{fig:flux-uncond} repeats the diagnostic on the two FLUX.2-klein-base
backbones. The unconditional distribution again contracts from a diverse base
onto a narrow hub (both backbones converge onto recurring portrait scenes
after NFT) and reopens under repair with varied subjects, styles, and
rendering modes, confirming that the contraction-and-reopening pattern, and
hence the hub-reading use of the unconditional route, is not specific to
SD3.5-M.

\section{Extended Comparisons}
\label{sec:extended-quant}

\begin{table*}[!t]
  \centering
  \renewcommand{\arraystretch}{1.08}
  \resizebox{\textwidth}{!}{%
  \begin{tabular}{lccccccc|cccc}
    \toprule
    & \multicolumn{7}{c|}{\textbf{PickScore Test Set}} &
    \multicolumn{4}{c}{\textbf{GenEval Test Set}} \\
    \cmidrule(lr){2-8} \cmidrule(lr){9-12}
    \textbf{Method} &
    PickScore $\uparrow$ & Aesthetic $\uparrow$ & ImageReward $\uparrow$ &
    HPSv2 $\uparrow$ & LPIPS-Div $\uparrow$ & DreamSim-Div $\uparrow$ &
    DINO-Div $\uparrow$ &
    GenEval $\uparrow$ & LPIPS-Div $\uparrow$ & DreamSim-Div $\uparrow$ &
    DINO-Div $\uparrow$ \\
    \midrule
    SD3.5-M &
      \gref{21.73} & \gref{6.026} & \gref{1.06} & \gref{0.295} &
      \gref{0.599} & \gref{0.242} & \gref{0.251} &
      \gref{0.654} & \gref{0.687} & \gref{0.339} & \gref{0.379} \\
    \midrule
    + DiffusionNFT &
      \cellcolor{graybg}23.51 & 6.592 & 1.44 & 0.327 &
      0.430 & 0.119 & 0.112 &
      \cellcolor{graybg}0.946 & 0.292 & 0.149 & 0.186 \\
    + E$^2$PO (reproduced) &
      \cellcolor{graybg}23.17 & 6.312 & 1.34 & 0.327 &
      0.498 & 0.166 & 0.159 &
      \cellcolor{graybg}0.921 & 0.463 & 0.204 & 0.258 \\
    + \textbf{Ours} &
      \cellcolor{graybg}23.26 & 6.344 & 1.41 & 0.323 &
      0.565 & 0.189 & 0.182 &
      \cellcolor{graybg}0.937 & 0.496 & 0.231 & 0.295 \\
    \bottomrule
  \end{tabular}%
  }
  \caption{\textbf{Comparison with Locally Reproduced E$^2$PO on SD3.5-M (PickScore and GenEval Test Sets).} All metrics are computed under the same
  evaluation
  protocol. Unlike the main paper, which reuses E$^2$PO's paper-reported
  reward values, this table uses our local reproduction throughout.
  \colorbox{graybg}{Shaded}: in-domain training reward.}
  \label{tab:e2po-merged}
\end{table*}

\begin{table*}[!t]
  \centering
  \renewcommand{\arraystretch}{1.08}
  \resizebox{0.9\textwidth}{!}{%
  \begin{tabular}{lccccccc}
    \toprule
    \textbf{Method} &
    PickScore $\uparrow$ & Aesthetic $\uparrow$ & ImageReward $\uparrow$ &
    HPSv2 $\uparrow$ & LPIPS-Div $\uparrow$ & DreamSim-Div $\uparrow$ &
    DINO-Div $\uparrow$ \\
    \midrule
    \multicolumn{8}{l}{\emph{FLUX.2-klein-base-4B}} \\
    Base &
      \gref{21.26} & \gref{5.995} & \gref{0.99} & \gref{0.289} &
      \gref{0.533} & \gref{0.246} & \gref{0.274} \\
    + DiffusionNFT &
      \cellcolor{graybg}23.38 & 6.723 & 1.35 & 0.320 &
      0.433 & 0.127 & 0.122 \\
    + \textbf{Ours} &
      \cellcolor{graybg}23.13 & 6.489 & 1.39 & 0.318 &
      0.485 & 0.169 & 0.157 \\
    \midrule
    \multicolumn{8}{l}{\emph{FLUX.2-klein-base-9B}} \\
    Base &
      \gref{22.04} & \gref{5.921} & \gref{1.24} & \gref{0.298} &
      \gref{0.526} & \gref{0.232} & \gref{0.253} \\
    + DiffusionNFT &
      \cellcolor{graybg}23.35 & 6.699 & 1.42 & 0.331 &
      0.425 & 0.130 & 0.126 \\
    + \textbf{Ours} &
      \cellcolor{graybg}23.04 & 6.442 & 1.45 & 0.322 &
      0.493 & 0.175 & 0.173 \\
    \bottomrule
  \end{tabular}%
  }
  \caption{\textbf{Quantitative Results on Two FLUX.2-klein-base Backbones (PickScore Test Set).} \colorbox{graybg}{Shaded}: in-domain training reward.
  Parenthesized rows: reference only, excluded from comparison.}
  \label{tab:flux}
\end{table*}

\subsection{Full Comparison with E$^2$PO}
\label{sec:e2po-full}

E$^2$PO~\cite{hu2026e2po} is not open-sourced, so its published checkpoint
and evaluation code
are unavailable for direct comparison. The main paper therefore reuses the
paper-reported reward values of E$^2$PO and omits diversity metrics. We have since
reproduced E$^2$PO locally, matching its official training budget and
following its published configuration as closely as possible;
Table~\ref{tab:e2po-merged} reports the full comparison using these
reproduced values throughout. The reproduced reward (23.17 PickScore, 0.921
GenEval) is slightly lower than the paper-reported values (23.38, 0.932)
used in the main paper, likely due to implementation differences. The main
paper conservatively retains these stronger
paper-reported values; Table~\ref{tab:e2po-merged} instead
evaluates all methods under one protocol.

Compared to the reproduced E$^2$PO, \textbf{ReNFT} achieves higher diversity
on all six diversity metrics across both protocols while also retaining more
reward (98.9\% vs.\ 98.6\% PickScore, 99.0\% vs.\ 97.4\% GenEval relative to
the NFT endpoint). The interface-level approach does recover meaningful
diversity, but at a larger reward cost and to a lesser extent than our
internal-route repair.

\subsection{Cross-Backbone Generalization}
\label{sec:flux}

Table~\ref{tab:flux} extends the evaluation to two FLUX.2-klein-base~\cite{flux2} backbones
(4B and 9B) on the PickScore test set. The NFT adapter of each backbone is trained
for 200 steps and repaired for another 50 steps, with the same
hyperparameters as on SD3.5-M except for the larger LoRA (rank 64, alpha
128); evaluation uses 20-step sampling, with the base model at CFG 4 and the
NFT and \textbf{ReNFT} checkpoints CFG-free. The collapse pattern reproduces
on both backbones: DiffusionNFT raises
PickScore by 2.12 and 1.31 points over the base model while DreamSim-Div
falls by 48\% and 44\%, respectively, matching the reward--diversity tension
on SD3.5-M. \textbf{ReNFT}
transfers without retuning: it retains 98.9\% (4B) and 98.7\% (9B) of
NFT's reward while improving DreamSim-Div by 33\% and 35\%, with consistent
gains on LPIPS-Div and DINO-Div as well. The repair mechanism therefore generalizes beyond
SD3.5-M: across a different backbone, LoRA capacity, training budget, and
evaluation protocol, the same internal routes exist and the same 50-step budget recalibrates them without retuning.

\section{Extended Qualitative Results}
\label{sec:extended-qual}
\begin{figure*}[!t]
  \centering
  \includegraphics[width=0.86\textwidth]{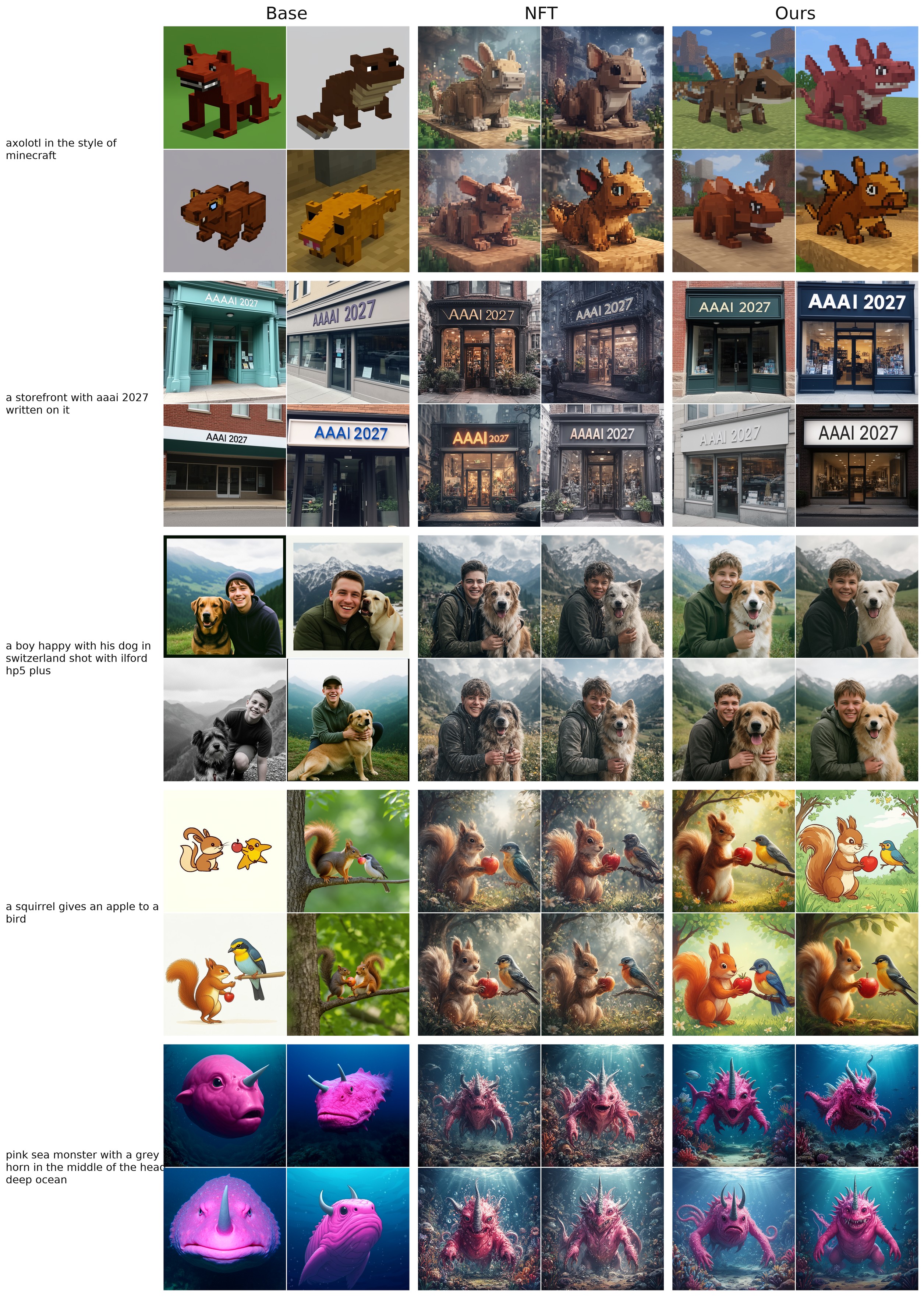}
  \caption{\textbf{Qualitative Comparison on FLUX.2-klein-base-4B.} Each row is one PickScore prompt; each cell shows four samples from the same method under different initial noises.}
  \label{fig:flux4b-cond}
\end{figure*}

\begin{figure*}[!t]
  \centering
  \includegraphics[width=0.86\textwidth]{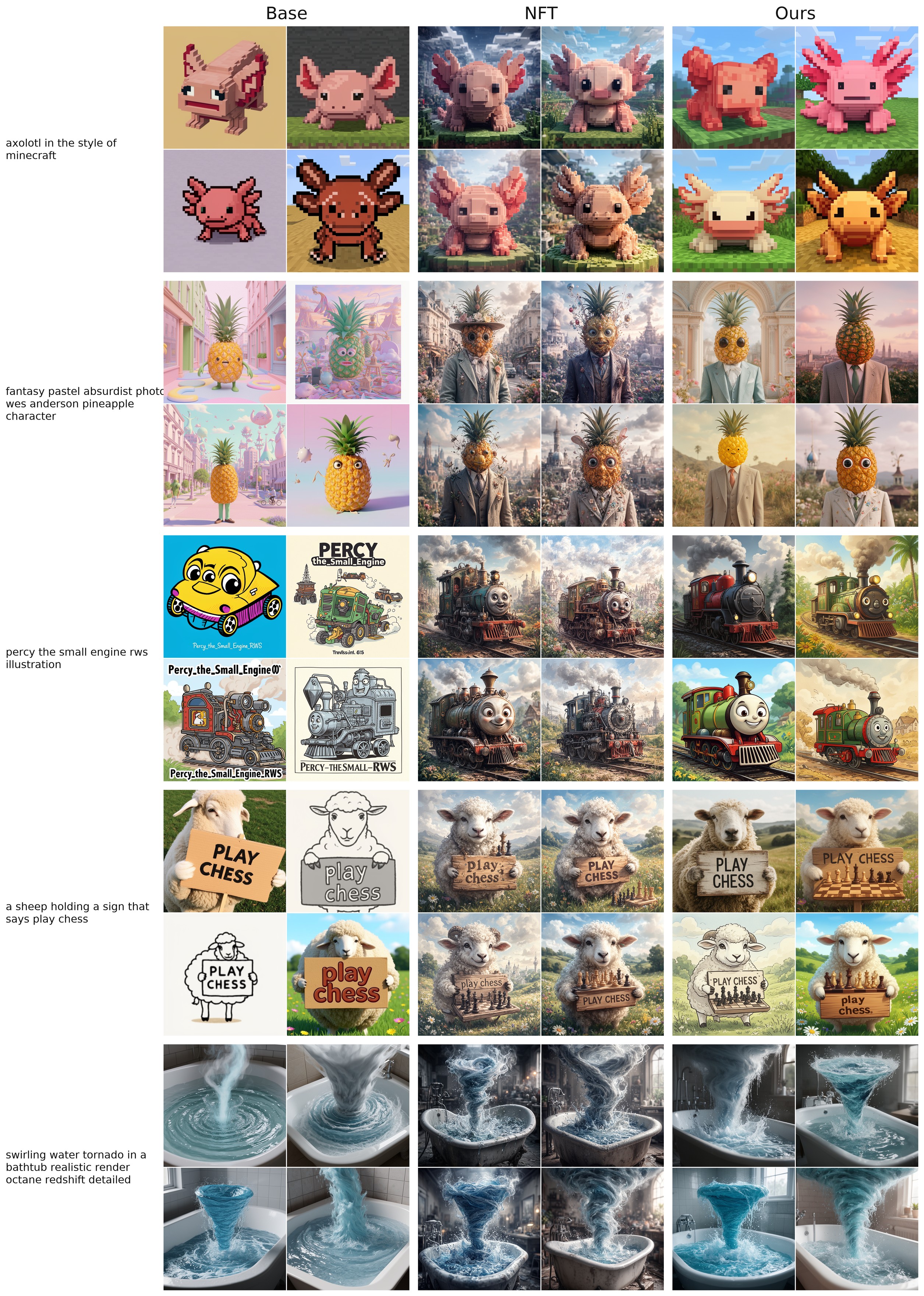}
  \caption{\textbf{Qualitative Comparison on FLUX.2-klein-base-9B.} Same layout as Figure~\ref{fig:flux4b-cond}.}
  \label{fig:flux9b-cond}
\end{figure*}

Figures~\ref{fig:flux4b-cond} and~\ref{fig:flux9b-cond} compare Base, NFT,
and \textbf{ReNFT} on five PickScore prompts per backbone, with four samples
per method under different initial noises. We discuss three representative
prompts per backbone below; the shared \emph{axolotl} prompt enables a
cross-backbone comparison at the end.

\noindent\textbf{FLUX.2-klein-base-4B.} On the \emph{axolotl-in-the-style-of-Minecraft}
prompt, all three methods adopt a side-view subject structure. Base has clean,
simple backgrounds; NFT fixes the structure across samples with visible noise
artifacts; \textbf{ReNFT} maintains the side-view structure but with fuller
backgrounds and textures than Base, without NFT's artifacts. On the
\emph{storefront-with-AAAI-2027} prompt, \textbf{ReNFT} achieves higher
text-spelling accuracy than both Base and NFT, with diversity approaching
Base and far exceeding NFT. On the \emph{squirrel-gives-an-apple-to-a-bird}
prompt, Base shows high diversity in composition and style, though some
samples have implausible object relationships; \textbf{ReNFT} produces more
structurally coherent scenes with style diversity markedly higher than NFT.

\noindent\textbf{FLUX.2-klein-base-9B.} On the same
\emph{axolotl-in-the-style-of-Minecraft} prompt, the 9B base adopts a
\emph{front-view} subject structure, distinct from the side view of the 4B
base, and NFT correspondingly collapses to a different hub, with
\textbf{ReNFT} reopening within the 9B's own range.
On the \emph{fantasy-pastel-Wes-Anderson-pineapple-character} prompt, Base
already produces prompt-relevant pineapple characters; both NFT and
\textbf{ReNFT} converge on pineapple-head with humanoid-body forms, likely
reflecting PickScore's preference for human-like subjects. \textbf{ReNFT}
has cleaner and more varied backgrounds compared to NFT's dense, ornate
settings. On the \emph{sheep-holding-a-sign-that-says-play-chess} prompt,
Base shows high diversity; NFT fixes to a narrow mode with only minor
structural differences across samples; \textbf{ReNFT} maintains multiple
compositional styles and subject structures.

\noindent\textbf{Cross-backbone axolotl comparison.} Under the same prompt,
the two backbones' Base, NFT, and \textbf{ReNFT} samples remain visually
distinct: 4B and 9B each concentrate on a different high-detail NFT hub and
reopen toward different backbone-specific ranges. This provides qualitative
evidence consistent with Section~4.1 of the main paper: post-training
reweights the inherited distribution of each model rather than introducing a shared
new mode (Figure~1 of the main paper). The backbone-specific reopening is also
consistent with the suppression-not-deletion hypothesis.

\noindent\textbf{Common NFT artifacts.} Beyond the per-prompt structural
collapse described above, the NFT columns in both grids share recurring
visual symptoms: a warm, oversaturated color palette; densely textured or
ornate backgrounds; and fine noise artifacts that give the images an
over-processed appearance. These symptoms are consistent across both
FLUX.2-klein-base backbones and match the hacking artifacts observed in the
SD3.5-M unconditional display (Section~3.2), suggesting a shared
reward-hacking signature rather than a backbone-specific artifact. These
visual artifacts (over-processed textures, unnatural detail, and palette
fixation) are not genuine quality improvements but shortcuts that
overfit the preferences of the reward model, exploiting evaluation blind spots
rather than producing subjectively higher-quality images.

Across the illustrated prompts, \textbf{ReNFT} is visibly more varied than
NFT in composition, color, and rendering style, while Table~\ref{tab:flux}
shows that this recovery retains most of NFT's reward.



\end{document}